%% file: main.tex
\documentclass{article}

\usepackage{eccv26}
\usepackage[utf8]{inputenc} 
\usepackage[T1]{fontenc}    
\usepackage{hyperref}       
\usepackage{url}            
\usepackage{booktabs}       
\usepackage{amsfonts}       
\usepackage{nicefrac}       
\usepackage{microtype}      
\usepackage{lipsum}
\usepackage{amsmath}
\usepackage{algorithm}
\usepackage{algpseudocode}
\usepackage{multirow}
\usepackage{fancyhdr}       
\usepackage{graphicx}       
\graphicspath{{media/}}     
\usepackage{caption}
\usepackage{subcaption}
\usepackage{listings}
\usepackage{microtype}

\title{N-Tree Diffusion for Long-Horizon Wildfire Risk Forecasting
}

\author{
  Yucheng Xing \\
  Department of Electrical and Computer Engineering \\
  Stony Brook University \\
  Stony Brook, NY 11794, USA \\
  \texttt{yucheng.xing@stonybrook.edu} \\
  \And
  Xin Wang \\
  Department of Electrical and Computer Engineering \\
  Stony Brook University \\
  Stony Brook, NY 11794, USA \\
  \texttt{x.wang@stonybrook.edu} \\
}

\begin{document}
\maketitle


\input{tex/section_0.tex}

\input{tex/section_1.tex}

\input{tex/section_2.tex}

\input{tex/section_3.tex}

\input{tex/section_4.tex}

\input{tex/section_5.tex}

\bibliographystyle{splncs04}
\bibliography{eccv26} 

\input{tex/append}

\end{document}

%% file: tex/section_0.tex


\begin{abstract} 

Long-horizon wildfire risk forecasting requires generating probabilistic spatial fields under sparse event supervision while maintaining computational efficiency across multiple prediction horizons. Extending diffusion models to multi-step forecasting typically repeats the denoising process independently for each horizon, leading to redundant computation. We introduce \textbf{\textit{N-Tree Diffusion (NT-Diffusion)}}, a hierarchical diffusion model designed for long-horizon wildfire risk forecasting. Fire occurrences are represented as continuous \textbf{\textit{Fire Risk Maps (FRMs)}}, which provide a smoothed spatial risk field suitable for probabilistic modeling. Instead of running separate diffusion trajectories for each predicted timestamp, NT-Diffusion shares early denoising stages and branches at later levels, allowing horizon-specific refinement while reducing redundant sampling. We evaluate the proposed framework on a newly collected real-world wildfire dataset constructed for long-horizon probabilistic prediction. Results indicate that NT-Diffusion achieves consistent accuracy improvements and reduced inference cost compared to baseline forecasting approaches. 
    
\end{abstract}

%% file: tex/section_1.tex
\section{Introduction~\label{sec:1}}


Wildfire risk forecasting is important in fire-prone regions, where early estimation of spatial risk can support response planning. From a modeling perspective, this task requires estimating spatial risk fields from sparse and highly imbalanced event observations. Fire occurrences are rare, spatially dispersed, and may emerge at arbitrary locations, making point-wise classification and fixed-location forecasting formulations insufficient. Moreover, long-horizon prediction involves generating risk estimates across multiple future timestamps, introducing additional challenges in uncertainty modeling and computational scalability. These characteristics motivate treating wildfire forecasting as a structured probabilistic spatial prediction problem rather than a binary detection task. 

Despite these modeling characteristics, most existing wildfire prediction approaches are formulated either as deterministic forecasting or binary classification tasks. When cast as a fire-or-not prediction problem, the rarity of fire events leads to severe class imbalance~\cite{zhang2019forest, zhang2021deep, zhang2024current, oak2024novel, nur2022creation, you2023pso}, making it difficult to learn stable decision boundaries. Deterministic forecasting models~\cite{cheng2008integrated, dutta2013deep, chavalithumrong2021learning, liang2019neural, natekar2021forest, kondylatos2022wildfire, li2023attentionfire_v1, cheng2022data,cheng2024deep, kadir2023wildfire, hu2024developing, gopu2023comparative, miao2023time, cao2024forest}, commonly used in other spatio-temporal prediction settings, typically operate on fixed locations or low-dimensional outputs and may not adequately represent spatially continuous risk across large geographic areas. In addition, most available datasets provide binary point annotations, which offer limited context. Although segmentation-based methods~\cite{hodges2019wildland, santopaolo2021forest, allaire2021emulation, ray2021predicting, li2023predicting, shadrin2024wildfire, jiang2023wfnet, marjani2024application} introduce spatial structure, they are generally designed for dense object regions and are less suited to scenarios where fire events appear as isolated and sparsely distributed points on a map. To model spatial uncertainty, diffusion-based generative models~\cite{ho2020denoising, song2020denoising} offer a natural mechanism for probabilistic spatial generation. However, when extended to multi-horizon forecasting, independently executing denoising trajectories for each timestamp introduces redundant computation and limits practical scalability.

In this work, we formulate long-horizon wildfire forecasting as the conditional generation of spatial risk fields. We introduce the \textit{Fire Risk Map (FRM)}, a continuous spatial representation that converts sparse fire annotations into smoothed probabilistic risk fields. FRM enables the modeling of uncertain and variable numbers of fire events without relying on binary labels and serves as a forecasting representation rather than a physics-based fire spread simulator. Building upon this formulation, we propose \textit{N-Tree Diffusion (NT-Diffusion)}, a hierarchical diffusion design tailored for long-horizon wildfire risk forecasting. Instead of executing independent diffusion processes for each prediction horizon, NT-Diffusion shares early denoising stages and branches at later levels through a tree-structured trajectory organization. This hierarchical sharing reduces duplicated sampling while preserving horizon-specific refinement. We further introduce a \textit{Shifting Diffusion} operation to align diffusion states across adjacent horizons and a \textit{Dual-Path Shifting Loss} to stabilize shared trajectory learning. We evaluate the proposed framework on a real-world wildfire dataset constructed for long-horizon probabilistic prediction. Experimental results show that NT-Diffusion improves forecasting accuracy while reducing inference costs compared to baseline approaches. Our contributions are summarized as follows:
\begin{itemize}
    \item We formulate long-horizon wildfire forecasting as the conditional generation of spatial risk fields and introduce the Fire Risk Map (FRM) for probabilistic representation under sparse supervision. 
    \item We propose NT-Diffusion, a hierarchical diffusion design that amortizes computation across prediction horizons through structured trajectory sharing. 
    \item We design a Shifting Diffusion operation and a Dual-Path Shifting Loss to support stable shared diffusion trajectory learning. 
    \item We provide an empirical evaluation on a real-world wildfire dataset, analyzing both forecasting performance and computational cost. 
\end{itemize}

The remainder of the paper is organized as follows: In Sec.~\ref{sec:2}, we briefly review the related works. Sec.~\ref{sec:3} presents the details of our proposed method. In Sec.~\ref{sec:4}, we evaluate its effectiveness through extensive experiments. Finally, Sec.~\ref{sec:5} concludes the paper with a discussion of limitations and future research directions. 

%% file: tex/section_2.tex
\section{Related Works~\label{sec:2}}


\input{tex/section_2.1}


%% file: tex/section_2.1.tex

Historically, wildfire prediction has been studied under statistical and machine learning frameworks that model fire risk using environmental and meteorological factors~\cite{li2022predictive, oliveira2021wildfire, jain2020review}. These approaches typically operate on pre-defined spatial locations and focus on point-level or regional risk estimation. 

With the advancement of deep learning, sequential models such as RNNs~\cite{cheng2008integrated, dutta2013deep, chavalithumrong2021learning}, LSTMs~\cite{liang2019neural, natekar2021forest, kondylatos2022wildfire, li2023attentionfire_v1, cheng2022data, cheng2024deep, kadir2023wildfire, hu2024developing, gopu2023comparative}, and Transformers~\cite{miao2023time, cao2024forest} have been widely applied to wildfire forecasting. These methods model temporal dependencies in historical observations and exogenous variables, but are generally formulated on fixed-location representations or aggregated regional features. While effective for location-specific forecasting, such formulations are not directly designed for spatially continuous, region-wide generative prediction.

To better capture spatial correlations, convolutional architectures have been adopted for fire classification~\cite{zhang2019forest, zhang2021deep, zhang2024current, oak2024novel, nur2022creation, you2023pso} and segmentation tasks~\cite{hodges2019wildland, santopaolo2021forest, allaire2021emulation, ray2021predicting, li2023predicting, shadrin2024wildfire, jiang2023wfnet, marjani2024application}. A plethora of methods~\cite{jin2020ufsp, huot2020deep, zhang2022dynamic, marjani2024cnn, li2021wildland, bhowmik2023multi, marjani2023firepred, burge2020convolutional, masrur2024capturing} have even attempted to combine both spatial and temporal information for fire forecasting. However, most available wildfire datasets provide either point-level fire occurrences or image-level binary annotations, limiting direct supervision for dense spatio-temporal risk field forecasting.

Beyond wildfire prediction, generative modeling~\cite{brooks2022generating, wang2022latent, bhagat2020disentangling, xie2020motion} has recently emerged as a powerful framework for high-dimensional spatio-temporal generation. Diffusion models~\cite{ho2020denoising, song2020denoising}, in particular, have demonstrated strong capability in modeling complex spatial distributions and long-horizon sequence generation. Several works~\cite{ho2022video, yu2023video} have explored diffusion-based modeling for video generation and spatio-temporal synthesis, as well as efficiency improvements through structured sampling strategies and trajectory reorganization. 

In contrast to prior wildfire forecasting approaches, we formulate long-horizon wildfire risk prediction as a conditional generative modeling problem over spatially continuous regional risk fields. Recent works have also begun exploring generative modeling in fire-related scenarios~\cite{zhou2025firesentry, rianto2025generative}, focusing on indoor environments or fine-grained short-term spread forecasting. Our work differs in both problem formulation and modeling strategy. We target long-horizon regional risk field generation and introduce a hierarchical N-Tree diffusion organization that enables structured trajectory sharing across forecasting horizons, reducing redundant computations while preserving generative expressiveness.

%% file: tex/section_3.tex
\section{Methodology~\label{sec:3}}


In this section, we first describe how long-horizon wildfire risk forecasting is formulated as the conditional generation of spatial risk fields through the construction of the \textit{Fire Risk Map (FRM)}. We then present \textit{N-Tree Diffusion (NT-Diffusion)}, a hierarchical diffusion design tailored for this forecasting setting, together with the \textit{Shifting Diffusion} mechanism for diffusion trajectory alignment across horizons. Finally, we introduce the \textit{Dual-Path Shifting Loss}, a training objective designed to support stable shared trajectory learning.

\subsection{Fire Risk Map Representation~\label{sec:3.1}}

Traditional fire forecasting methods are typically formulated under a fixed-location assumption. In one common formulation, a pre-defined set of $K$ spatial coordinates (e.g., selected facilities or grid points) is specified in advance, and the model predicts a vector of length $K$ representing fire probability at each of these locations. Another class of approaches adopts a keypoint-style detection framework, where the model outputs $K$ feature maps corresponding to $K$ potential fire locations, and the final coordinates are obtained by identifying local maxima within each map. Although these formulations differ in implementation, they both require the number of predicted fire locations to be fixed in advance. As a result, the modeling capacity is constrained by a pre-defined hyperparameter $K$, which limits flexibility when the number and spatial distribution of fire events vary across time and regions. 

To avoid this limitation, we represent fire occurrence risk using a continuous \textit{Fire Risk Map (FRM)}, which provides a spatially dense risk field over the entire geographic region. This representation removes the need to pre-define a fixed number of key points and allows the model to naturally adapt to varying numbers and spatial distributions of fire events across different scenarios, without restricting the output dimensionality to a fixed set of candidate locations. In addition, FRM supports flexible downstream decision-making. Instead of committing to a hard fire/no-fire threshold during forecasting, it provides a continuous risk surface that can be thresholded or aggregated according to application-specific criteria (e.g., regional conditions or operational requirements). 

\begin{figure}[!htpb]
    \centering
    \begin{subfigure}{.24\linewidth}
        \centering
        \includegraphics[width=\linewidth]{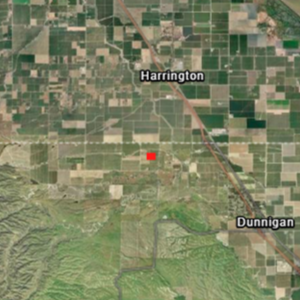}
        \caption{}
    \end{subfigure}
    \hfill
    \begin{subfigure}{.24\linewidth}
        \centering
        \includegraphics[width=\linewidth]{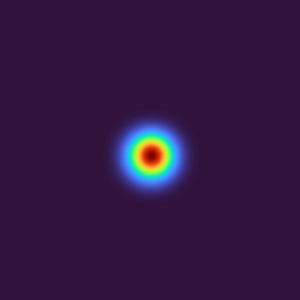}
        \caption{}
    \end{subfigure}
    \hfill
    \begin{subfigure}{.24\linewidth}
        \centering
        \includegraphics[width=\linewidth]{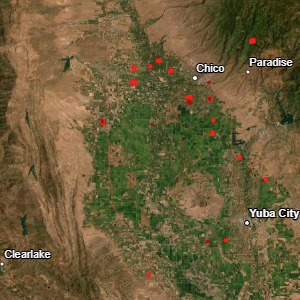}
        \caption{}
    \end{subfigure}
    \hfill
    \begin{subfigure}{.24\linewidth}
        \centering
        \includegraphics[width=\linewidth]{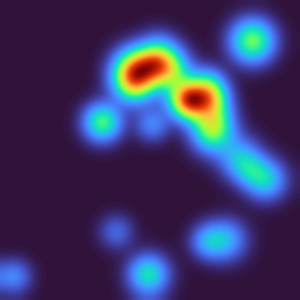}
        \caption{}
    \end{subfigure}
    \caption{Illustration of the Fire Risk Map (FRM).(a) shows a single fire occurrence in the area and (b) presents its Gaussian risk representation. For multiple real fire events occurring in the region corresponding to the real map, as depicted in (c), the resulting FRM is generated as shown in (d).~\label{fig:fpm}}
\end{figure}

Since the true risk of real-world events, such as a fire, is inherently difficult to measure, most available datasets only record indicators denoting whether a fire event occurred or not at a specific location. To construct a spatially continuous representation from such discrete observations, we transform isolated fire records into localized spatial kernels. This smoothing converts discrete annotations into a continuous spatial risk representation. Specifically, as shown in Fig.~\ref{fig:fpm}, for each recorded fire located at coordinate $\mu_{k, t} = (x_{k, t}, y_{k, t})$ at time $t$, we construct a Gaussian kernel centered at this location:
\begin{equation}
    p_{k, t}(x, y) = 
    \frac{1}{2\pi\sigma_{x, k, t}\sigma_{y, k, t}}\exp{(-\frac{(x - x_{k, t})^2}{2\sigma_{x, k, t}^2} - \frac{(y - y_{k, t})^2}{2\sigma_{y, k, t}^2})}.~\label{eq:gaussian_1}
\end{equation}
The bandwidth parameters $\sigma_{x, k, t}$ and $\sigma_{y, k, t}$ are determined based on fire intensity measurements (e.g., brightness and temperature observed by satellites) associated with each fire event. This intensity-dependent smoothing modulates the spatial spread of each kernel according to observed fire intensity, resulting in a representation that reflects both spatial uncertainty and event magnitude. 

The FRM at time $t$ is constructed by aggregating all Gaussian kernels within the region:
\begin{equation}
    I_{t}(x, y) = \sum_{k = 1}^{K_{t}}p_{k, t}(x, y),~\label{eq:fpm} 
\end{equation}
where $K_{t}$ denotes the number of recorded fire events at time $t$. The resulting map $I_{t}$ provides a spatially continuous representation suitable for conditional generative modeling of future wildfire risk fields. 

\noindent\textbf{Notation:} Our objective is to generate a sequence of future fire risk maps $\{I_{0, 0}, I_{1, 0}, \dots, I_{T, 0}\}$ over $T+1$ forecasting horizons. The generative process is implemented using a diffusion-based model with $D$ steps. We index diffusion states by noise level $s$, where $s = D$ is the most noisy state and $s = 0$ is the clean map. The reverse diffusion process updates $I_{t, s} \to I_{t, s - 1}$. The following subsections describe how these maps are generated through the proposed model architecture.

\subsection{N-Tree Diffusion Generation Structure~\label{sec:3.2}}

To generate a sequence of future fire risk maps, we employ conditional diffusion models as a generative backbone for spatial risk field prediction. Diffusion models provide a flexible framework for modeling high-dimensional spatial outputs under conditioning signals. In a direct multi-horizon formulation, a full reverse diffusion process of $D$ steps can be applied independently to each forecasting horizon. Generating $T+1$ horizons in this manner requires $(T+1)\times D$ diffusion steps, leading to computational costs that scale linearly with the number of prediction horizons. 

\begin{figure*}[!htpb]
    \centering
    \includegraphics[width=\linewidth]{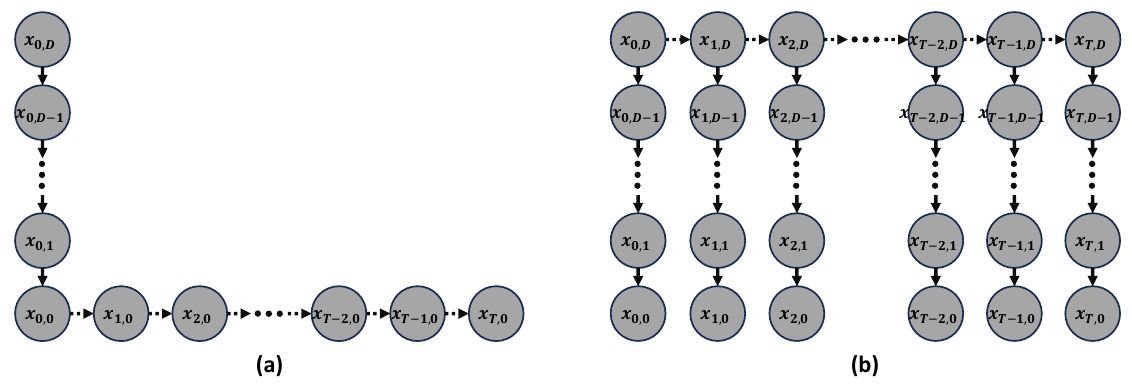}
    \caption{Two extreme computational structures for multi-horizon diffusion generation. (a) Fully shared reverse trajectory. (b) Fully independent reverse trajectories.~\label{fig:framework1}}
\end{figure*}

More generally, multi-horizon diffusion generation can be organized in different computational structures. One extreme formulation shares a single diffusion trajectory across all forecasting horizons (Fig.~\ref{fig:framework1} (a)). In this setting, the reverse diffusion process is executed once to obtain a base output, and subsequent horizons are produced through lightweight temporal transitions applied at the output level. This formulation minimizes the computation since only one trajectory of length $D$ is required, but it limits horizon-specific refinement within the diffusion process itself. At the other extreme, diffusion trajectories are fully separated across horizons (Fig.~\ref{fig:framework1} (b)), allowing complete horizon-specific modeling at the expense of increased computational cost. 

These two formulations reflect different allocations of modeling capacity across horizons. Diffusion models, however, exhibit a structured reverse evolution governed by progressively decreasing noise levels. Early reverse stages operate under high noise and primarily establish low-frequency global structure. As the reverse process proceeds and noise diminishes, later stages enable increasingly high-resolution spatial refinement. In other words, representational resolution increases along the reverse trajectory. This staged progression implies that different segments of the reverse process operate at different structural scales. 
In early reverse stages, the representation remains dominated by high noise levels and carries limited horizon-specific information. Under this regime, multiple horizons can share common reverse states without significantly restricting expressiveness, as fine-grained horizon-dependent variations emerge primarily in later low-noise stages.

\begin{figure*}[!htpb]
    \centering
    \includegraphics[width=\linewidth]{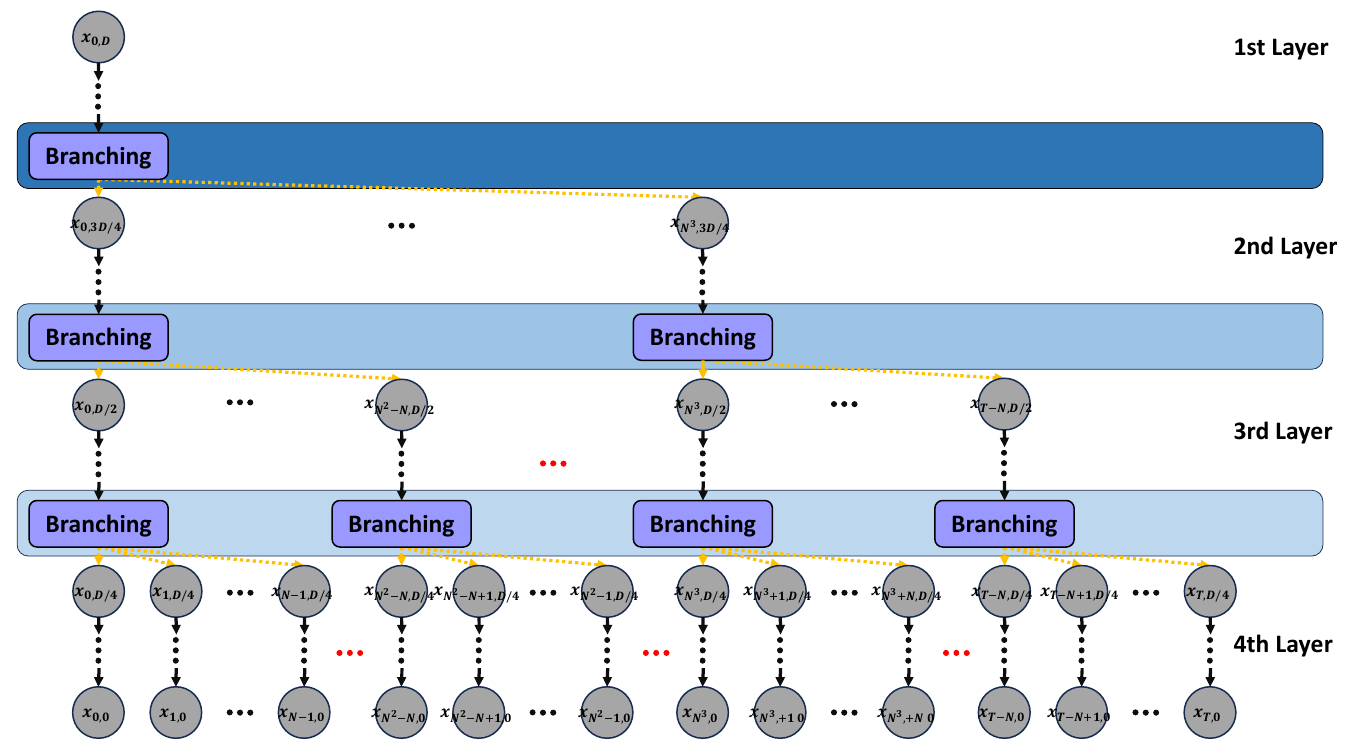}
    \caption{Illustration of NT-Diffusion with $L=4$ hierarchical layers, where reverse trajectories are partially shared and progressively branched across horizons.}~\label{fig:nt_diff}
\end{figure*}

Based on this structural property of diffusion models, we propose \textit{N-Tree Diffusion (NT-Diffusion)}, a hierarchical organization of the reverse process that enables partial trajectory sharing across forecasting horizons, as shown in Fig.~\ref{fig:nt_diff}. Rather than fully sharing or fully separating diffusion trajectories, NT-Diffusion partitions the reverse process into multiple segments and introduces controlled branching between segments. Specifically, the total number of reverse steps $D$ is divided into $L$ equal segments (referred to as hierarchical layers in Fig.~\ref{fig:nt_diff}), each consisting of $D / L$ steps. Generation begins from a fully noisy representation $I_{0,D}$. After completing the first segment of $D/L$ reverse steps, the intermediate representation is branched into $N$ child trajectories using the proposed \textit{Shifting Diffusion} operation (described in Sec.~\ref{sec:3.3}). Each branch then proceeds independently through the next segment before further branching. Recursively applying this procedure constructs a tree-structured hierarchy with depth $L$ and branching factor $N$. 

Under this construction, the number of leaf nodes (i.e., final forecasting horizons) satisfies 
\begin{equation}
    N^{L-1} = T + 1, ~\label{eq:NLT} 
\end{equation}
and the total number of reverse steps in NT-Diffusion is therefore
\begin{equation}
    \begin{aligned}
        S_{\text{NT-Diff}} &= 1 \times \frac{D}{L} + N \times \frac{D}{L} + \dots + N^{L-1} \times \frac{D}{L} \\
                        &= (1 + N + \dots + N^{L-1}) \times \frac{D}{L} \\
                        &= \frac{N^{L} - 1}{N - 1} \times \frac{D}{L}. ~\label{eq:diff_step_our}
    \end{aligned}
\end{equation}
Compared to the fully independent formulation, whose steps are
\begin{equation}
    S_{\text{traditional}} = (T + 1) \times D,~\label{eq:diff_step_1b}
\end{equation}
the computational reduction factor becomes
\begin{equation}
    R = \frac{S_{\text{traditional}}}{S_{\text{NT-Diff}}} = \frac{N^{L-1} \times D}{\frac{N^{L} - 1}{N - 1} \times \frac{D}{L}} = \frac{L \times (N^{L} - N^{L-1})}{N^{L} - 1} = \frac{L\times(N-1)\times(T + 1)}{N\times(T+1) - 1},~\label{eq:speedup}
\end{equation}
where we use the relation implied by Eq.~\ref{eq:NLT}.
This expression makes explicit how the computational reduction depends on the tree organization. Since $D$ and $T$ are fixed in a given forecasting setup, the tree parameters are coupled by Eq.~\ref{eq:NLT}. The reduction factor can therefore be viewed as a function of the tree depth $L$, with different choices of $L$ inducing different sharing schedules along the reverse diffusion process. A detailed analysis is provided in Appendix.~\ref{sec:computation_analysis}.

\subsection{Branching with Shifting Diffusion Mechanism~\label{sec:3.3}}

In NT-Diffusion, multiple forecasting horizons are generated by branching from shared intermediate diffusion states. At a branching node, a single parent representation is reused to initialize several child trajectories corresponding to different horizons. Without additional differentiation, these child trajectories would receive identical inputs at the branching point, which may limit the model's ability to produce horizon-specific variations in subsequent reverse diffusion updates. To enable horizon-specific divergence after branching, we introduce a \textit{Shifting Diffusion} mechanism that injects relative horizon information into the reverse diffusion update when a shared parent state is expanded into multiple child horizons. 

Let $I_{t, s}$ denote the noisy state at forecasting horizon $t$ and noise level $s$. During standard (non-branching) diffusion updates, each horizon evolves from its own previous state, so the horizon index remains unchanged. At a branching point, however, multiple horizons share a common parent state originating from horizon $t_{p}$. In this case, the model must distinguish between child horizons $t$ that originate from the same parent representation. We therefore encode the relative horizon offset $\Delta t = t - t_{p}$ and provide it as an additional conditioning signal to the model. This relative index acts as a horizon identifier for branching transitions. It primarily serves as a horizon identifier that differentiates child trajectories originating from the same shared diffusion state.

\begin{figure}[!htpb]
    \centering
    \includegraphics[width=0.65\linewidth]{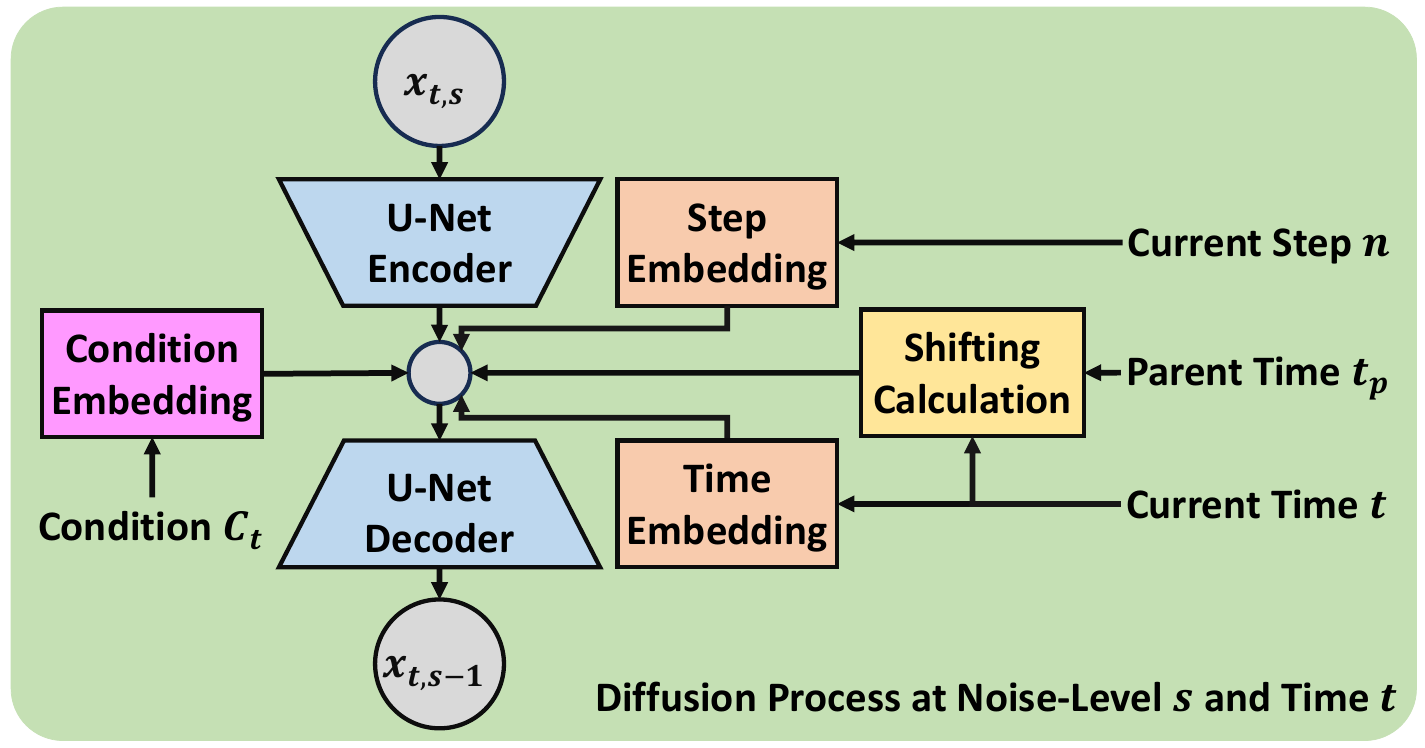}
    \caption{Illustration of the diffusion denoising process at noise level $s$ and time $t$.~\label{fig:shifting_diff}}
\end{figure}

The Shifting Diffusion module, illustrated in Fig.~\ref{fig:shifting_diff}, is implemented using the following components:

\begin{itemize}
    \item[$\diamond$] \textbf{U-Net backbone:} A U-Net architecture~\cite{ronneberger2015u} serves as the main denoising model, consisting of an encoder $G_{\text{E}}(\cdot)$ and a decoder $G_{\text{D}}(\cdot)$. The encoder maps the noisy input $I_{t, s}$ into latent features, and the decoder predicts the updated state $I_{t, s - 1}$.
    \item[$\diamond$] \textbf{Condition embedding:} The generation process is conditioned on available contextual information denoted as $C_{t}$, associated with forecasting horizon $t$. Depending on data availability, $C_{t}$ may include observed historical fire states, static geographic features (e.g., terrain or land cover), and, when available, auxiliary exogenous signals such as meteorological forecasts. A condition encoder $F_{\text{CE}}$ extracts feature representations $h_{\text{cond}} = F_{\text{CE}}(C_{t})$, which guide the generation of future horizon-dependent risk maps.
    \item[$\diamond$] \textbf{Step embedding:} The current noise level $s$ is encoded using positional encoding~\cite{vaswani2017attention} to produce a step embedding $h_{\text{step}} = F_{\text{SE}}(s)$, allowing the model to adapt to varying noise magnitudes.
    \item[$\diamond$] \textbf{Horizon and shift embedding:} The absolute forecasting horizon index $t$ is encoded as $h_{\text{time}} = F_{\text{TE}}(t)$, using the same positional encoding technique to preserve the ordering information. In addition, the relative offset $\Delta t = t - t_{p}$ is encoded as a shift embedding $h_{\text{shift}} = F_{\text{Shift}}(\Delta t)$, where $F_{\text{Shift}}(\cdot)$ shares the same structure as $F_{\text{SE}}(\cdot)$ and $F_{\text{TE}}(\cdot)$. During non-branching steps, $t = t_{p}$, so that $\Delta t = 0$. At branching steps, $\Delta t \neq 0$, enabling distinct modulation for different child horizons. 
\end{itemize} 

The latent feature is modulated using FiLM conditioning~\cite{perez2018film}:
\begin{equation}
    I_{t, s - 1} = G_{\text{D}}(\gamma \odot G_{\text{E}}(I_{t, s}) + \beta), ~\label{eq:unet_film}
\end{equation}
where
\begin{equation}
    \gamma, \beta = \text{Linear}\left([h_{\text{cond}}, h_{\text{step}}, h_{\text{time}}, h_{\text{shift}}]\right).~\label{eq:film}
\end{equation}

Through this design, shared reverse diffusion segments maintain common intermediate representations across horizons, while branching transitions introduce horizon-specific modulation through explicit relative index encoding. This mechanism allows NT-Diffusion to combine structured sharing with controlled divergence across forecasting horizons.

\subsection{Dual-Path Shifting Loss~\label{sec:3.4}}

In our model, the additional learnable components beyond the standard diffusion backbone reside in the Shifting Diffusion mechanism, which is responsible for differentiating child trajectories at branching points while preserving the denoising capability of the base diffusion process. To simultaneously maintain standard diffusion behavior and enable effective branching transitions, we introduce a \textit{Dual-Path Shifting Loss (DPSL)}. 

Given a pair of fire risk maps $\{I_{t_{i}, 0}, I_{t_{j}, 0}\}$ and a pre-defined noise schedule $\{\beta_{1}, \dots, \beta_{D}\}$, a diffusion step index $d \in [1, D]$ is uniformly sampled during training. A noisy state is generated through the standard forward diffusion process:
\begin{equation}
    I_{t_{i}, s} = \sqrt{\overline{\alpha}_{s}}I_{t_{i}, 0} + \sqrt{1 - \overline{\alpha}_{s}}\epsilon,~\label{eq:diffusion_forward_1}
\end{equation}
where
\begin{equation}
    \overline{\alpha}_{s} = \prod_{k = 1}^{s}(1 - \beta_{k}).~\label{eq:diffusion_forward_2}
\end{equation}
and $\epsilon \sim \mathcal{N}(0, I)$. To preserve the generative capability of conventional diffusion models, we adopt the standard DDPM objective: 
\begin{equation}
    \begin{aligned}
        \mathcal{L}_{1} &= \mathcal{L}(\frac{I_{t_{i}, s} - \sqrt{\overline{\alpha}_{s}}I_{t_{i}, 0}}{\sqrt{1 - \overline{\alpha}_{s}}} - \epsilon_{\theta}(I_{t_{i}, s}, C_{t_{i}}, s, t_{i}, t_{i} - t_{i})) \\
        &= \mathcal{L}(\epsilon - \epsilon_{\theta}(I_{t_{i}, s}, C_{t_{i}}, s, t_{i}, 0)), ~\label{eq:loss_1}
    \end{aligned}
\end{equation}
where $\epsilon_{\theta}(\cdot)$ denotes the Shifting Diffusion module and $\mathcal{L}(\cdot)$ is the weighted DDPM loss formulation~\cite{ho2020denoising}. This term ensures that when no branching occurs (i.e., $\Delta t = 0$), the model behaves identically to a standard diffusion model. 

Meanwhile, to enable effective branching transitions, we introduce a complementary objective that enables horizon-specific differentiation from shared noisy states under relative horizon shifts. Instead of performing a separate forward diffusion process for $I_{t_{j}, 0}$, we reuse the noisy state $I_{t_{i}, s}$ and condition the model on a nonzero relative shift $\Delta t = t_{j} - t_{i}$. The objective is defined as
\begin{equation}
    \mathcal{L}_{2} = \mathcal{L}(\frac{I_{t_{i}, s} - \sqrt{\overline{\alpha}_{s}}I_{t_{j}, 0}}{\sqrt{1 - \overline{\alpha}_{s}}} - \epsilon_{\theta}(I_{t_{i}, s}, C_{t_{j}}, s, t_{j}, t_{j} - t_{i})). ~\label{eq:loss_2}
\end{equation}
This auxiliary term does not alter the forward diffusion formulation. Rather, it encourages the model to learn horizon-dependent transformations from a shared noisy state by injecting relative horizon information through the shift embedding. Since $\mathcal{L}_{2}$ shares the same functional structure as the standard DDPM objective and operates under the same noise schedule, it preserves the denoising objective while introducing structured cross-horizon supervision. 

The overall DPSL is defined as 
\begin{equation}
    \mathcal{L}_{DPSL} = \mathcal{L}_1 + \mathcal{L}_2.~\label{eq:DPSL}
\end{equation}
During paired training, symmetric constraints are applied in both directions for a sampled pair $(t_{i}, t_{j})$, resulting in four structurally identical loss terms. These terms share the same scale and formulation, so we adopt uniform weighting to avoid introducing additional hyperparameters and to maintain stable optimization. Through DPSL, NT-Diffusion preserves standard diffusion denoising behavior along non-branching paths while explicitly supervising horizon differentiation at branching transitions. This dual-path design enables structured sharing across horizons without altering the underlying diffusion training formulation.

%% file: tex/section_4.tex
\section{Experiments~\label{sec:4}}


\subsection{Experimental Setting~\label{sec:4.1}}

\subsubsection{Datasets~\label{sec:4.1.1}} 

Existing wildfire datasets are typically designed for alternative prediction settings, such as location-specific fire occurrence modeling~\cite{yavas_2025_14712845, singla2021wildfiredb} or whole-image binary fire/no-fire classification~\cite{kaggle_wildfire_dataset}. Some datasets further focus on fine-scale fire spread analysis using high-resolution imagery. However, these formulations are not designed for long-horizon generative forecasting of spatially continuous regional risk fields. To align with the proposed forecasting formulation in Sec.~\ref{sec:3.1}, we process satellite-based fire observations into a spatio-temporal risk field representation. Specifically, we collect daily fire locations across the continental United States recorded by the Moderate Resolution Imaging Spectroradiometer (MODIS)~\cite{MODIS} between Apr. 30, 2015 and Apr. 29, 2025, from the NASA Fire Information for Resource Management System (FIRMS)~\cite{nasaNASAFIRMS}. The raw FIRMS data provide fire coordinates and observation timestamps. Following Sec.~\ref{sec:3.1}, daily fire records are converted into spatially continuous FRMs using Gaussian kernel aggregation. The resulting maps are normalized to $[0,1]$ and resized to $128 \times 128$. The dataset contains $3653$ daily maps. We construct temporal segments of length $28$ days, where the first frame is used as the conditioning input and the subsequent $27$ frames correspond to future forecasting horizons. To avoid temporal leakage, training, validation, and testing splits are defined chronologically at the day level before segment construction. Temporal segments are generated within each split independently, ensuring that no frame appears in more than one split. The final split ratio is $70\% : 15\% : 15\%$.

\subsubsection{Evaluation Metrics~\label{sec:4.1.2}}

We evaluate both forecasting quality and computational efficiency. For forecasting quality, predicted FRMs are compared with the target FRMs constructed from recorded fire coordinates following Sec.~\ref{sec:3.1}. We adopt pixel-level metrics, including Root Mean Square Error (RMSE, $\downarrow$), Mean Absolute Error (MAE, $\downarrow$), and KL-Divergence (KL, $\downarrow$), computed between normalized predicted and target spatial risk fields. Since our objective is continuous risk field forecasting rather than binary decision-making, threshold-dependent classification metrics such as accuracy or recall are not reported. For computational efficiency, we report inference time (Time\,[ms], $\downarrow$), total floating-point operations (FLOPs\,[G], $\downarrow$), and the number of model parameters (Params\,[M], $\downarrow$).

\subsubsection{Implementation Details~\label{sec:4.1.3}}

All models are trained from scratch on a single NVIDIA RTX A6000 GPU (48GB). 
The AdamW optimizer is used with a cosine annealing learning rate schedule. 
The learning rate is set from $1e-3$ to $1e-5$ for non-diffusion baselines and from $1e-4$ to $1e-6$ for diffusion-based models. To ensure a fair comparison, we use a batch size of $4$ during training for all models and set the batch size to $1$ during inference. For NT-Diffusion, the tree depth is set to $L=4$. The diffusion process uses $D_{\text{train}}=1000$ steps during training and $D_{\text{test}}=10$ steps during inference. The U-Net backbone consists of three residual blocks in both the encoder and decoder, one middle residual block, and a final convolutional layer, forming a $15$-layer architecture. The SiLU activation function is adopted throughout the network. The implementation details are provided to facilitate reproducibility. 

\subsection{Overall Performance~\label{sec:4.2}}

To evaluate the performance of our model, we compare NT-Diffusion with representative methods for wildfire forecasting. To align all baselines with the FRM formulation, minor architectural adjustments are applied where necessary to accommodate dense spatial inputs. Pure temporal models (e.g.,~\cite{natekar2021forest, kadir2023wildfire, cao2024forest}) operate on pre-defined locations and vectorized inputs. To ensure compatibility with spatially dense FRMs, these models are reformulated to operate on pixel-level representations. All baselines are trained and evaluated on identical data splits under consistent experimental settings. As shown in Table.~\ref{tab:overall}, NT-Diffusion achieves the best forecasting accuracy across all quality metrics. The improvement is consistent across RMSE, MAE, and KL divergence, indicating more accurate reconstruction of continuous risk fields. In terms of model complexity, NT-Diffusion maintains a compact parameter size compared to several temporal baselines. In particular, vector-based recurrent formulations tend to incur larger parameter counts when extended to dense spatial representations, whereas NT-Diffusion directly models spatial fields with a comparatively small number of parameters. Regarding inference latency, diffusion-based generation incurs higher runtime than lightweight recurrent models. However, within the diffusion framework, the proposed N-Tree structure reduces computational cost compared to fully independent diffusion generation, as analyzed in Sec.~\ref{sec:4.3}.
\begin{table}[!htpb]
    \centering
    \renewcommand\arraystretch{0.8}
    \caption{Overall performance comparison among different fire forecasting models.~\label{tab:overall}}
    \begin{tabular}{c c c c c c c}
        \toprule
            &\multicolumn{3}{c}{Quality ($\times$1e-2, $\downarrow$)} &\multicolumn{3}{c}{Efficiency ($\downarrow$)} \\
        \cmidrule(lr){2-4}\cmidrule(lr){5-7}
            & RMSE & MAE & KL & Time & FLOPs &Params \\ 
        \midrule
            LSTM~\cite{kadir2023wildfire} &6.15$\pm$0.13 &2.10$\pm$0.01 &5.58$\pm$0.25 &2.12$\pm$0.17 &36.24 &1342.32 \\
            ULSTM~\cite{bhowmik2023multi} &6.85$\pm$1.89 &1.84$\pm$0.96 &6.32$\pm$2.33 &2.54$\pm$0.02 &5.30 &26.51 \\ 
            CNN~\cite{santopaolo2021forest} &6.23$\pm$0.49 &2.25$\pm$0.93 &5.68$\pm$0.45 &47.16$\pm$0.04 &174.16 &3.49 \\ 
            ConvLSTM~\cite{kondylatos2022wildfire} &11.49$\pm$1.47 &4.76$\pm$1.60 &19.73$\pm$12.19 &7.75$\pm$0.10 &9.77 &1.02 \\ 
            Transformer~\cite{cao2024forest} &6.35$\pm$0.01 &2.30$\pm$0.08 &6.64$\pm$0.24 &8.13$\pm$0.09 &466.68 &253.97 \\
            ViT~\cite{he2024deep} &6.01$\pm$0.04 &1.47$\pm$0.08 &5.18$\pm$0.03 &8.82$\pm$0.19 &8.55 &30.51 \\ 
        \midrule
            \textbf{NT-Diffusion} &\textbf{5.86$\pm$0.04} &\textbf{1.16$\pm$0.01} &\textbf{5.16$\pm$0.05} &30.35$\pm$0.22 &18.36 &6.23 \\ 
        \bottomrule
    \end{tabular}
\end{table}

\subsection{Ablation Study~\label{sec:4.3}}

To evaluate the effectiveness of the proposed N-Tree diffusion structure in terms of forecasting accuracy and computational efficiency, we compare NT-Diffusion with several commonly used diffusion architectures, including: a) \textbf{Diffusion (2D-AR)}, where each frame is processed by a 2D-UNet and generated auto-regressively from the previous frame; b) \textbf{Diffusion (2D-Attn)}, where each frame is processed by a 2D-UNet in parallel with temporal attention applied across frames after each diffusion step; c) \textbf{Diffusion (3D)}, where the entire frame sequence is processed by a 3D-UNet; 
and d) \textbf{Diffusion (ViT)}, where the sequence is modeled using a ViT-based architecture. All models are trained with $D_{\text{train}}=1000$ steps and evaluated with $D_{\text{test}}=10$ steps. As shown in Table.~\ref{tab:diffusion}, NT-Diffusion achieves the lowest RMSE and KL divergence among the compared diffusion models, while maintaining competitive MAE. At the same time, it significantly reduces inference time and FLOPs compared to alternative diffusion structures. These results demonstrate that structured trajectory sharing improves computational efficiency without degrading forecasting accuracy.

\begin{table}[!htpb]
    \centering
    \renewcommand\arraystretch{0.8}
    \caption{Comparison among different diffusion-based models (steps=10, mean$\pm$std, \textbf{Bold}: best, \underline{Underline}: second best).~\label{tab:diffusion}}
    \begin{tabular}{c c c c c c c}
        \toprule
            &\multicolumn{3}{c}{Quality ($\times$1e-2, $\downarrow$)} &\multicolumn{3}{c}{Efficiency ($\downarrow$)} \\
        \cmidrule(lr){2-4}\cmidrule(lr){5-7}
            & RMSE & MAE & KL & Time & FLOPs &Params  \\ 
        \midrule
            Diffusion(2D-AR) &\underline{5.87$\pm$0.01} &\textbf{1.13$\pm$0.00} &\underline{5.17$\pm$0.01} &727.62$\pm$2.95 &\underline{32.42} &\textbf{5.89} \\ 
            Diffusion(2D-Attn) &5.94$\pm$0.03 &\underline{1.16$\pm$0.01} &5.27$\pm$0.04 &\underline{32.18$\pm$0.24} &34.60 &14.29 \\  
            Diffusion(3D) &5.94$\pm$0.02 &1.15$\pm$0.01 &5.27$\pm$0.03 &99.05$\pm$0.39 &83.67 &12.60 \\ 
            Diffusion(ViT) &5.95$\pm$0.04 &1.25$\pm$0.02 &5.30$\pm$0.06 &94.75$\pm$0.29 &63.64 &38.88 \\ 
        \midrule
            \textbf{NT-Diffusion} &\textbf{5.86$\pm$0.04} &\underline{1.16$\pm$0.01} &\textbf{5.16$\pm$0.05} &\textbf{30.35$\pm$0.22} &\textbf{18.36} &\underline{6.23} \\ 
        \bottomrule
    \end{tabular}
\end{table}

\begin{figure*}[!htpb]
    \centering
    \includegraphics[width=\linewidth]{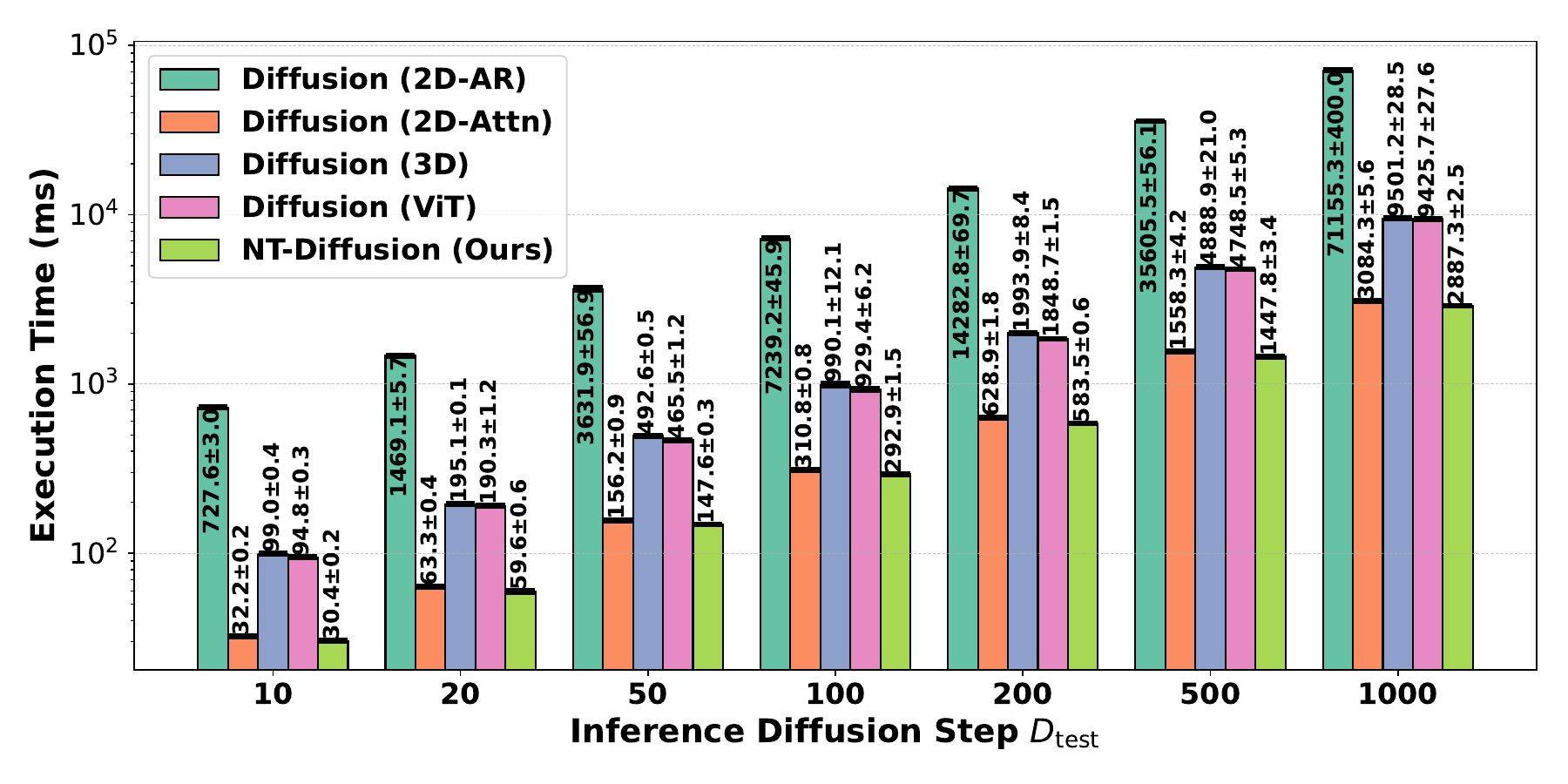}
    \caption{Comparison of execution time (ms) among diffusion-based models with different steps.~\label{fig:diffusion_time}}
\end{figure*}

While Table.~\ref{tab:diffusion} reports results under a fixed sampling step ($D_{\text{test}}=10$), it does not fully reflect how computational cost scales with the number of diffusion iterations. To further analyze the scalability of different diffusion structures, we evaluate their execution time under varying testing steps $D_{\text{test}}$, as shown in Fig.~\ref{fig:diffusion_time} (numerical values are provided in Table.~\ref{tab:append_diffusion_time} within Appendix.~\ref{sec:time_table}). Across all models, inference time increases approximately linearly with the number of sampling steps, consistent with the iterative nature of diffusion generation. NT-Diffusion consistently achieves the lowest execution time across all tested settings. Although the relative speed-up remains stable, the absolute time gap increases as $D_{\text{test}}$ grows. This behavior indicates that NT-Diffusion maintains a consistently lower computational cost per diffusion step, resulting in increasing absolute time savings for longer diffusion trajectories.

\begin{figure*}[!htpb]
    \centering
    \includegraphics[width=\linewidth]{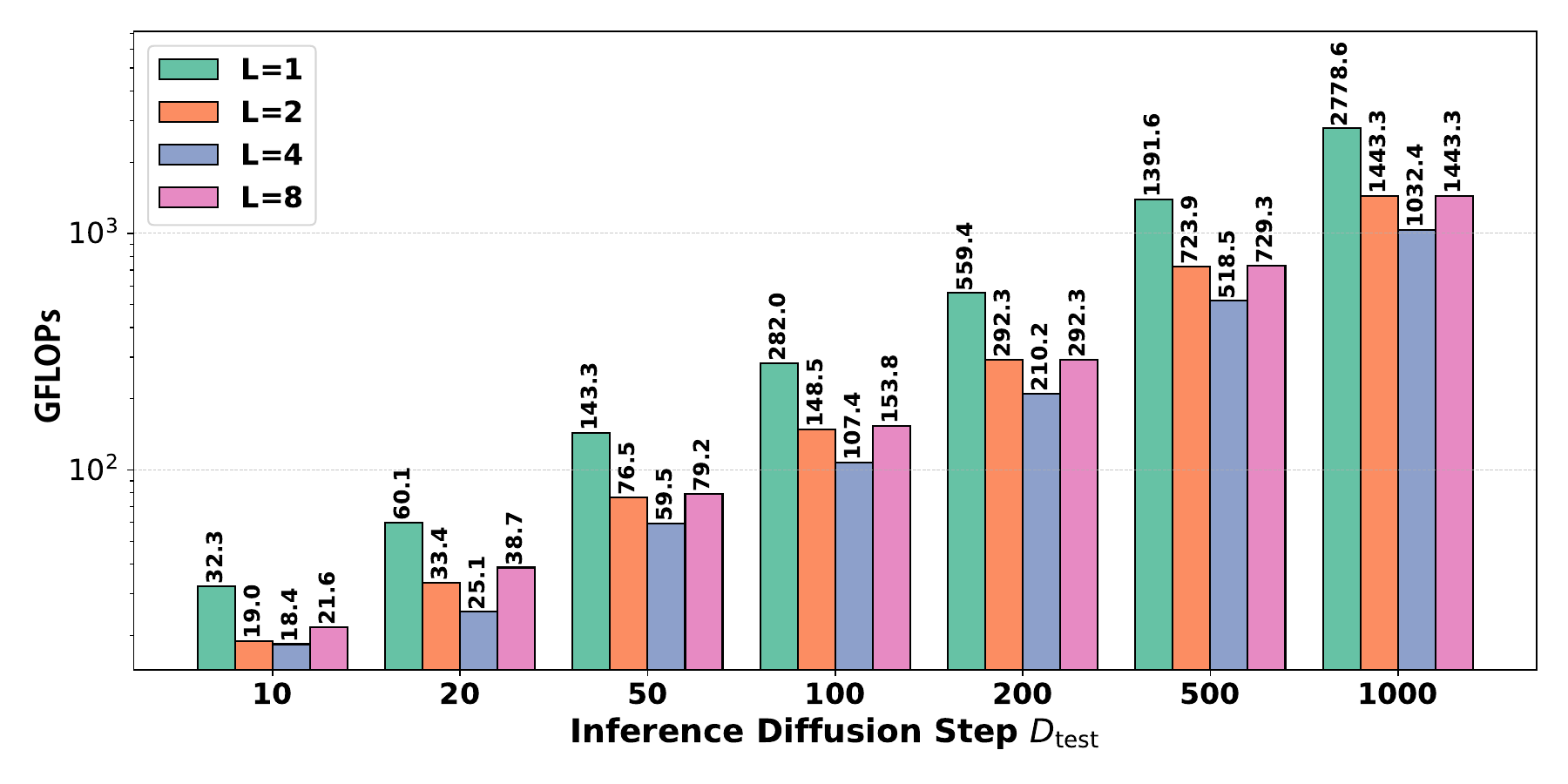}
    \caption{Comparison of average FLOPs of our model with different hyperparameter $L$.~\label{fig:L_flops}}
\end{figure*}

We further analyze the effect of different tree depths $L$ on computational cost, as shown in Fig.~\ref{fig:L_flops} (execution time are provided in Table.~\ref{tab:L_time} within Appendix.~\ref{sec:time_table}). Consistent with the theoretical analysis in Sec.~\ref{sec:3.2}, Increasing $L$ initially reduces total computations by allocating a larger portion of the reverse diffusion trajectory to shared segments across horizons. However, since the tree parameters are constrained by the structural relation in Eq.~\ref{eq:NLT}, both $L$ and the induced branching factor $N$ are discrete and coupled. As a result, the computational cost does not decrease monotonically for all $L$. Under the current forecasting horizon configuration, $L=4$ achieves the lowest FLOPs, representing a favorable balance between sharing depth and branching overhead. Further increasing $L$ is constrained by the discrete tree structure in Eq.~\ref{eq:NLT}, which limits additional sharing benefits and results in slightly increased computations.

%% file: tex/section_5.tex
\section{Conclusion~\label{sec:5}}


In this paper, we formulate long-horizon wildfire forecasting as a conditional generative modeling problem over spatially continuous risk fields. To address the computational challenges of multi-horizon diffusion generation, we propose NT-Diffusion, a hierarchical N-Tree diffusion framework that enables structured trajectory sharing across forecasting horizons. The proposed Shifting Diffusion mechanism further supports horizon differentiation within shared reverse processes through explicit relative index modulation. By organizing reverse diffusion trajectories in a tree-structured manner, our framework reduces redundant computations while preserving generative capacity. Experimental results on a real-world satellite-based wildfire dataset demonstrate consistent improvements in forecasting accuracy, parameter efficiency, and diffusion scalability. Future work includes exploring adaptive branching strategies and extending the framework to broader spatio-temporal generative forecasting scenarios. We emphasize that the proposed model is intended to complement existing expert-driven monitoring systems and should be integrated into decision-support pipelines rather than deployed as a standalone automated solution.

%% file: tex/append.tex
\newpage
\appendix
\renewcommand\thesubsection{\thesection\arabic{subsection}}
\section*{Appendix}

\input{tex/complexity_analysis}

\input{tex/efficiency_table}

\input{tex/section_2.2}

%% file: tex/complexity_analysis.tex
\subsection{Computational Reduction Analysis~\label{sec:computation_analysis}}

We analyze how the computational reduction factor depends on the tree structure parameters when the forecasting horizon $T$ and diffusion trajectory length $D$ are fixed. Recall that the total number of reverse steps in NT-Diffusion is 
\begin{equation}
    S_{\text{NT-Diff}} = \frac{N^{L} - 1}{N - 1} \times \frac{D}{L},\label{eq:app1}
\end{equation}
and the fully independent strategy requires
\begin{equation}
    S_{\text{traditional}} = (T + 1) \times D.\label{eq:app2}
\end{equation}
The reduction factor is therefore 
\begin{equation}
    R = \frac{S_{\text{traditional}}}{S_{\text{NT-Diff}}}.\label{eq:app3}
\end{equation}
Under the structural constraint of the tree, 
\begin{equation}
    N^{L-1} = T+1,\label{eq:app4}
\end{equation}
which implies 
\begin{equation}
    N = (T + 1)^{\frac{1}{L - 1}}.\label{eq:app5}
\end{equation}
Using the relation $N^{L} = N \times (T + 1)$ from Eq.~\ref{eq:app4}, the reduction factor in Eq.~\ref{eq:app3} can be written exactly as
\begin{equation}
    R = \frac{L\times(N-1)\times(T + 1)}{N\times(T+1) - 1}.\label{eq:app6}
\end{equation}
Substituting $N$ according to Eq.~\ref{eq:app5}, we obtain an explicit expression in terms of the tree depth $L$:
\begin{equation}
    R(L) = \frac{L\times(T + 1)\times\left((T + 1)^{\frac{1}{L - 1}} - 1\right)}{(T + 1)^{\frac{L}{L - 1}} - 1}.\label{eq:app7}
\end{equation}
This expression characterizes precisely how the computational reduction depends on the chosen tree depth $L$ when the total number of forecasting horizons $T + 1$ is fixed. 

For large $T$, the denominator in Eq.~\ref{eq:app6} satisfies 
\begin{equation}
    N\times (T + 1) - 1 \approx N \times (T + 1),\label{eq:app8}
\end{equation}
leading to the approximation
\begin{equation}
    R \approx L \times \frac{N - 1}{N} = L\times (1 - \frac{1}{N}),
\end{equation}
which provides intuition on how the depths $L$ and the induced branching factor $N$ jointly influence the achievable reduction. Increasing the depth $L$ shortens each reverse segment to $D/L$ and therefore introduces branching earlier along the reverse trajectory, while simultaneously reducing the branching factor $N$ under the constraint given in Eq.~\ref{eq:app4}. Consequently, deeper trees distribute branching across more layers with smaller branching factors, whereas shallower trees require larger branching factors applied over fewer layers. The exact expression above quantifies the computational reduction achieved under these different structural configurations.

%% file: tex/efficiency_table.tex
\subsection{Execution Time Comparison~\label{sec:time_table}}

For greater clarity and intuitiveness, we present the experimental data from Fig.~\ref{fig:diffusion_time} and Fig.~\ref{fig:L_flops} in the following tables.

\begin{table*}[!htpb]
    \centering
    \caption{Comparison of average execution time (ms, mean$\pm$std) among diffusion-based models with different diffusion steps $D_{test}$ (\textbf{Bold}: best).~\label{tab:append_diffusion_time}}
    \begin{tabular}{r r r r r r}
        \toprule
            \multirow{2}{*}{$D_{test}$} &\multicolumn{4}{c}{NT-Diffusion} &\multirow{2}{*}{\textbf{NT-Diffusion}} \\
        \cmidrule(lr){2-5}
             &\multicolumn{1}{c}{2D-AR} &\multicolumn{1}{c}{2D-Attn} &\multicolumn{1}{c}{3D} &\multicolumn{1}{c}{ViT}  \\
        \midrule
            10 &727.62$\pm$2.95 &32.18$\pm$0.24 &99.05$\pm$0.39 &94.75$\pm$0.29 &\textbf{30.35$\pm$0.22} \\
            20 &1469.10$\pm$5.65 &63.32$\pm$0.37 &195.13$\pm$0.14 &190.27$\pm$1.24 &\textbf{59.58$\pm$0.55} \\
            50 &3631.94$\pm$56.94 &156.24$\pm$0.88 &492.57$\pm$0.47 &465.52$\pm$1.16 &\textbf{147.63$\pm$0.28} \\
            100 &7239.19$\pm$45.91 &310.76$\pm$0.80 &990.10$\pm$12.14 &929.37$\pm$6.20 &\textbf{292.92$\pm$1.50} \\
            200 &14282.82$\pm$69.74 &628.90$\pm$1.82 &1993.92$\pm$8.39 &1848.71$\pm$1.48 &\textbf{583.50$\pm$0.59} \\
            500 &35605.53$\pm$56.11 &1558.31$\pm$4.16 &4888.90$\pm$20.99 &4748.53$\pm$5.34 &\textbf{1447.84$\pm$3.40} \\
            1000 &71155.30$\pm$400.04 &3084.30$\pm$5.62 &9501.18$\pm$28.52 &9425.70$\pm$27.61 &\textbf{2887.30$\pm$2.45} \\
        \bottomrule
    \end{tabular}
\end{table*}

\begin{table*}[!htpb]
    \centering
    \caption{Comparison of average execution time (ms, mean$\pm$std) of our model with different hyperparameter $L$.~\label{tab:L_time}}
    \begin{tabular}{r r r r r}
        \toprule
            \multirow{2}{*}{$D_{test}$} &\multicolumn{4}{c}{NT-Diffusion} \\
        \cmidrule(lr){2-5}
            &\multicolumn{1}{c}{$L=1$} &\multicolumn{1}{c}{$L=2$} &\multicolumn{1}{c}{$L=4$} &\multicolumn{1}{c}{$L=8$} \\
        \midrule
            10 &30.95$\pm$1.08 &30.44$\pm$0.29 &30.35$\pm$0.22 &30.94$\pm$0.36 \\
            20 &60.70$\pm$0.30 &60.21$\pm$0.42 &59.58$\pm$0.55 &60.66$\pm$0.33 \\
            50 &151.18$\pm$1.95 &149.64$\pm$2.40 &147.63$\pm$0.28 &149.41$\pm$2.73 \\
            100 &296.95$\pm$5.00 &295.43$\pm$0.38 &292.92$\pm$1.50 &296.16$\pm$2.73 \\
            200 &592.16$\pm$2.00 &586.79$\pm$1.61 &583.50$\pm$0.59 &591.32$\pm$4.10 \\
            500 &1464.09$\pm$1.22 &1432.50$\pm$0.94 &1447.84$\pm$3.40 &1513.95$\pm$1.28 \\
            1000 &2911.44$\pm$15.76 &2915.65$\pm$0.52 &2887.30$\pm$2.45 &2870.75$\pm$4.97 \\
        \bottomrule
    \end{tabular}
\end{table*}

%% file: tex/section_2.2.tex
\subsection{Background -- Diffusion Model for Image Sequence Generation~\label{sec:2.2}}

Image sequence generation focuses on synthesizing temporally coherent image sequences by learning the underlying data distribution. Most early approaches rely on GAN-based frameworks~\cite{brooks2022generating,clark2019adversarial,saito2017temporal,tulyakov2018mocogan,vondrick2016generating,wang2020g3an,wang2022latent} or are built upon VAE-based models~\cite{bhagat2020disentangling,denton2017unsupervised,li2018disentangled,xie2020motion}. These methods typically model spatial and temporal dependencies jointly, but often suffer from training instability or limited sample diversity.

More recently, diffusion models~\cite{ho2020denoising} have emerged as a powerful alternative for high-dimensional generative modeling. VDM~\cite{ho2022video} extends diffusion models to spatio-temporal settings by integrating temporal operations into the denoising backbone. Specifically, VDM replaces standard 2D convolutions in the U-Net~\cite{ronneberger2015u} with 3D convolutions to jointly capture spatial and temporal dependencies within each reverse diffusion step. While effective for sequence generation, operating directly in pixel space introduces substantial computational and memory overhead, which limits scalability to higher resolutions or longer sequences. To alleviate this issue, PVDM~\cite{yu2023video} performs diffusion in a low-dimensional latent space, significantly reducing computational cost while preserving generation quality.

Beyond direct generation of full sequences, several works explore structured decomposition strategies. VIDM~\cite{mei2023vidm} separates content and motion modeling, generating a reference frame first and then synthesizing motion conditioned on it. Similar decomposition strategies are adopted in LEO~\cite{wang2023leo} and LaMD~\cite{hu2023lamd}, where spatial and temporal components are modeled using different diffusion backbones. GD-VDM~\cite{lapid2023gd} further introduces multi-stage diffusion pipelines, generating intermediate representations before refining them through additional diffusion processes.

In addition to convolutional architectures, transformer-based backbones such as Vision Transformers (ViTs)~\cite{dosovitskiy2020image} have been incorporated into sequence diffusion models, as in VDT~\cite{lu2023vdt} and W.A.L.T.~\cite{gupta2023photorealistic}, enabling patch-based spatio-temporal modeling across entire sequences. While these approaches enhance representational capacity and temporal coherence, the computational cost of diffusion-based sequence generation remains substantial. In most existing frameworks, a full reverse diffusion trajectory is executed for each frame or for the entire sequence without explicitly exploiting redundancy across correlated outputs.

In diffusion-based sequence generation, early reverse diffusion stages typically capture coarse global structure and exhibit high similarity across neighboring horizons. Motivated by this, we introduce a structured trajectory-sharing mechanism that merges portions of diffusion paths across correlated outputs, reducing redundant computations while preserving horizon-specific refinements.